\def\year{2021}\relax
\documentclass[letterpaper]{article} % DO NOT CHANGE THIS
\usepackage{aaai21}  % DO NOT CHANGE THIS
\usepackage{times}  % DO NOT CHANGE THIS
\usepackage{helvet} % DO NOT CHANGE THIS
\usepackage{courier}  % DO NOT CHANGE THIS
\usepackage[hyphens]{url}  % DO NOT CHANGE THIS
\usepackage{graphicx} % DO NOT CHANGE THIS
\usepackage{natbib}  % DO NOT CHANGE THIS AND DO NOT ADD ANY OPTIONS TO IT
\usepackage{caption} % DO NOT CHANGE THIS AND DO NOT ADD ANY OPTIONS TO IT
\usepackage{multirow}

\usepackage{amsmath}
\usepackage{amsfonts}
\usepackage{amssymb} 
\DeclareMathOperator{\IoU}{IoU} 
\DeclareMathOperator{\mIoU}{mIoU} 
\DeclareMathOperator*{\argmax}{argmax}
\newtheorem{theorem}{Theorem}
\newtheorem{corollary}{Corollary}[theorem]
\usepackage[dvipsnames]{xcolor}
\usepackage{soul}
\title{Distribution-aware Margin Calibration for Medical Image Segmentation}

\author{
	Zhibin Li, 
    Litao Yu, 
    Jian Zhang
    \\
}
\affiliations{
   \textsuperscript{\rm 1}School of Electrical and Data Engineering\\
   University of Technology Sydney\\
   \{zhibin.li, litao.yu, jian.zhang\}@uts.edu.au
}

\begin{document}

\maketitle

\begin{abstract}
The Jaccard index, also known as Intersection-over-Union (IoU score), is one of the most critical evaluation metrics in medical image segmentation. However, directly optimizing the mean IoU (mIoU) score over multiple objective classes is an open problem. Although some algorithms have been proposed to optimize its surrogates, there is no guarantee provided for their generalization ability. In this paper, we present a novel data-distribution-aware margin calibration method for a better generalization of the mIoU over the whole data-distribution, underpinned by a rigid lower bound. This scheme ensures a better segmentation performance in terms of IoU scores in practice. We evaluate the effectiveness of the proposed margin calibration method on two medical image segmentation datasets, showing substantial improvements of IoU scores over other learning schemes using deep segmentation models. 

\end{abstract}

\section{Introduction}
The medical image segmentation is a critical yet challenging learning problem in medical data analysis. The task is to build a computation model to accurately locate and identify the region of interest, such as lesions and instruments in medical images, which can be used for automatic medical instrument control and related disease diagnosis followed by some proper treatments. Specifically, during the global pandemic COVID-19 recently, the segmentation of the infection lesions from Computed Tomography (CT) scans is very important for quantitative measurement of the disease progression for accurate diagnosis and the follow-up treatment. 

Recently, the development of deep convolutional neural networks (CNNs) has led to remarkable progress in image segmentation due to their powerful feature representation ability to describe the local visual properties. For deep learning-based image segmentation, the encoder-decoder like convolutional segmentation models, such as U-net \cite{MICCAI15:UNET} and its variants \cite{TMI:COPLENET,TMI:UNETPLUS}, can well handle the visual-semantic consistencies and have achieved very promising results. 

To train a reliable deep learning model for medical image segmentation, the learning objective function is one of the most critical ingredients. The most straightforward way is to treat the image segmentation as a dense classification task, which examines each pixel in images individually, comparing the class-predictions to the one-hot encoded target vector. Thus, categorical cross-entropy becomes the most intuitive loss function. The minimization of overall cross-entropy is directly related to the maximization of pixel accuracy. In the training process of deep segmentation models, cross-entropy loss averages over all pixels in images, which is essentially asserting equal learning to each pixel in an image batch. This can be problematic in medical image segmentation if the actual classes are in the imbalanced representation in the image corpus, as training can be dominated by the most prevalent class, e.g., the small foreground interest regions are submerged by large background areas. Although applying a cost-sensitive re-weighting scheme \cite{MICCAI19:ELL} to alleviate the data imbalance and emphasize the ``important'' pixels, it is unclear how to determine the weights for the best IoU scores. Furthermore, the measure of cross-entropy on the validation set is often a poor indicator of the model quality, because minimizing the pixel-wise loss cannot guarantee that the model can obtain a higher Jaccard index (or IoU score, Intersection-over-Union) or dice coefficient, which are more commonly used in image segmentation and can better sketch the contours of interest regions. To deal with this problem, some recently proposed loss functions have been proposed, e.g., IoU loss \cite{ISVC16:IOU}, dice loss \cite{TMI:DICE} and Focal-Tversky loss \cite{ISBI19:FOCAL_TVK}. However, these loss functions mainly aim to minimize the empirical IoU on the training dataset, which usually leads to over-fitting. The generalized performance, i.e., the expected IoU on the unknown test dataset, has not been investigated and cannot be guaranteed.

A ``better'' machine learning model should feature a better generalized performance, i.e., the performance measured on the underlying data distribution, where the testing instances are sampled from. Clearly, there is a gap between the empirical performance on the training dataset and the generalized performance. This gap is commonly called the error bound. Thus, optimizing the generalized performance can be achieved through (1) optimizing the empirical performance approximated by a surrogate loss associated with the performance metric, e.g. IoU; and (2) controlling the error bound through regularization terms such as $l_2$-norm or a weight-decay scheme. As such, in medical image segmentation tasks, learning a model towards the optimal Jaccard index, or mIoU, should also consider these two factors. However, to the best of our knowledge, no proper method has been designed for controlling the error bound directly related to mIoU optimization, which is rather critical for a better generalization of the model. The error bound regarding accuracy can be controlled by the margins among multiple classes, which is well known for its use in Support Vector Machines (SVMs) \cite{boser1992training}. For data-imbalanced learning problems, uneven margins can be applied to well calibrate the importance of specific classes to attend \cite{li2002perceptron,khan2019striking,cao2019learning}. In medical image segmentation, class imbalance largely exists in various datasets, which hinders the maximization of mIoU. Although Li et. al proposed a margin tuning method \cite{li2019overfitting}, their method is limited to binary segmentation and the parameter setting is empirical, which highly relies on the manual trials on different datasets. The power of the ``uneven'' margins inspire us to develop a proper margin calibration scheme, to control the error bound for the performance improvement in medical image segmentation.

In this paper, we propose a novel distribution-aware margin calibration method, to optimize the mIoU in medical image segmentation. The margins across multiple classes are pre-computed based on the label distribution, which can well calibrate the distance between foreground and background pixels in the loss computation. Our method has the following three compelling advantages over other learning objectives: (1) it provides a lower bound for data-distribution-mIoU, which means the model has a guaranteed generalization ability; (2) the margin-offsets can be efficiently computed, which is readily pluggable into deep segmentation models; (3) the proposed learning objective is directly related to IoU scores, i.e., it is consistent with the evaluation metric. Due to the high discriminative power and stability, it is worth using the proposed margin calibration method as a learning objective in the challenging medical image segmentation tasks. We conduct comprehensive experiments on two medical image segmentation datasets, which indicates our method is able to achieve a considerable improvement compared to other training objectives. 

\section{Related work}

Deep learning-based image segmentation models have achieved significant progress on large-scale benchmark datasets \cite{CVPR17:ADE20K,CVPR16:CITYSCAPES} in recent years. The deep segmentation methods can be generally divided into two streams: the fully-convolutional networks (FCNs) and the encoder-decoder structures. The FCNs \cite{CVPR15:FCN} are mainly designed for general segmentation tasks, such as scene parsing and instance segmentation. Most FCNs are based on a stem-network (e.g., Inception network \cite{AAAI17:INCEPTIONV4}) pre-trained on a large-scale dataset, and the dilated convolution is used to enlarge the receptive field for more contextual information. In the encoder-decoder structure \cite{TPAMI:SEGNET,3DV:VNET}, the encoder maps the original images into low-resolution feature representations, while the decoder mainly restores the spatial information with skip-connections. Such networks are usually light-weighted with fewer parameters, which have been extensively used for medical image segmentation. Combining the encoder-decoder structure and dilated convolution can effectively boost the pixel-wise prediction accuracy \cite{TPAMI:DEEPLAB}, but is extremely computational demanding.

As a dense prediction task, medical image segmentation aims to train light-weight models on comparably small datasets, to accurately sketch the contours of the region of interest, such as tumour and body organs. Thus, U-net \cite{MICCAI15:UNET} models and their variants are the best choices. Since the commonly used cross-entropy in classification cannot well reflect the segmentation quality in medical images, a better optimization objective should be well designed. A ``better'' loss function should be consistent with the evaluation metrics and discriminative to target class labels. In recent years, various loss functions have been proposed specifically for medical image segmentation, and most of them can be used in a plug-and-play way. For example, the distribution-based loss functions (e.g., weighted cross-entropy loss \cite{MICCAI15:UNET} and focal loss \cite{CVPR17:FOCAL}), the region-based loss functions (e.g., IoU loss \cite{ISVC16:IOU}, dice loss \cite{TMI:DICE} and Tversky loss \cite{MLMIW17:TVERSKY}) and boundary-based loss functions (e.g., Hausdorff distance loss \cite{TMI:HAUSDORFF} and Boundary loss \cite{MIDL19:BOUNDARY}). These loss functions can also be jointly used in model optimization \cite{ISBI19:FOCAL_TVK}. Applying either distribution-based loss or region-based loss functions, there exists a problem that the continuous class probability of each pixel is indirectly related to IoU scores. To deal with this problem, Maxim et al. proposed to use submodular measures to readily optimize the segmentation model in the continuous setting \cite{CVPR18:LOVASZ_SOFTMAX}. However, the above loss functions specifically designed for image segmentation mainly aim to minimize the empirical risk in the model training procedure, without the consideration of the generalization error over the underlying data distribution. In our work, we design a new margin calibration scheme to overcome this difficulty from the perspective of data-distribution-related error bound, which provides a better learning objective for medical image segmentation compared to other learning metrics, both theoretically and practically.    

\section{Our method} \label{SEC:METHOD}

\subsection{Problem setup and notations}

Image segmentation can be considered as a dense prediction learning task. Considering an input space $\mathcal{X}\in\mathbb{R}^m$, and the target space $\mathcal{Y} =\{1,...,c\}^m$, where $m$ is the number of image pixels. The function $\theta\in\Theta: \mathcal{X} \mapsto \mathbb{R}^{m\times c}$ is a complex non-linear projection from raw images to scores for all pixels regarding all classes. In deep learning-based methods, $\Theta$ can be a deep learning model with trainable parameters. Given an image $\mathbf{x}\in\mathcal{X}$ with a corresponding mask $\mathbf{y}\in\mathcal{Y}$, we denote the output score for $i$-th pixel regarding the $j$-th foreground class by $\theta_{ij}(\mathbf{x})$, and the predicted label is given by $\hat y_{i}=\argmax_{j\in[c]} \theta_{ij}(\mathbf{x})$. Then, given a vector of ground truth $\mathbf{y}$ and a predicted label vector $\mathbf{\hat y}$, the empirical IoU for class $k$ over an image of $m$ pixels is defined as:
\begin{equation}\label{EQ:IOU_1}
\IoU_{k,m}(\theta) = \frac{\sum\limits_{i=1}^{m} \mathbb{I}(y_i=k\wedge\hat{y}_i= k) }{\sum\limits_{i=1}^{m} \mathbb{I}(y_i=k \vee \hat{y}_i= k)},
\end{equation}
where $\mathbb{I}(\cdot)$ is an indicator function. It gives the ratio in [0, 1] of the intersection between the ground truth and the predicted mask over their union, with the convention that 0/0 = 1 \cite{CVPR18:LOVASZ_SOFTMAX}.

Let $p_{k0,m}(\theta)$ be the empirical probability that a foreground class $k$ pixel is observed but is predicted as the background class by $\theta$, i.e., $p_{k0,m}(\theta)=\frac{1}{m}\sum\limits_{i=1}^{m} \mathbb{I}(y_i=k \wedge\hat y_i\neq k)$. Similarly, $p_{0k,m}(\theta)$ denotes the empirical probability that a pixel of the background class is observed but is predicted as a foreground class $k$. We use $p_{k,m}$ to denote the empirical probability that a class $k$ pixel is observed, i.e. $p_{k,m} = \frac{1}{m}\sum\limits_{i=1}^{m}\mathbb{I}(y_i=k)$. So Eq. (\ref{EQ:IOU_1}) can be re-formulated by:

\begin{equation} \label{emiou}
\IoU_{k,m}(\theta) = \frac{p_{k,m} - p_{k0,m}(\theta)}{ p_{k,m} + p_{0k,m}(\theta)}.
\end{equation}

When there are $c$ classes presented, the empirical mean IoU (mIoU) is defined as $\mIoU_{m}(\theta) = \frac{1}{c} \sum\limits_{k=1}^{c} \IoU_{k,m}(\theta)$.

In the evaluation of the segmentation performance, IoU or mIoU is computed globally over an image dataset $D$, which contains $n$ pixels in total. Replacing $p_{k,m}$, $p_{k0,m}(\theta)$ and $p_{0k,m}$ in Eq.(\ref{emiou}) with $p_{k,n}$, $p_{k0,n}(\theta)$ and $p_{0k,n}(\theta)$, respectively, we can get the IoU and mIoU on the dataset $D$. We denote the output score of $i$-th pixel in a dataset $D$ regarding class $j$ by $s_{ij}(\theta, D)$, and denote its label by $y_i$. We use $s_{ij}$ to denote $s_{ij}(\theta, D)$ whenever there is no ambiguity.

We assume that the images in the dataset $D$ are independently and identically distributed (i.i.d) according to some unknown distribution $\mathcal{D}$ over $\mathcal{X}\times \mathcal{Y}$, and let $\mathcal{D}_{\mathcal{Y}}$ denote the projection of $\mathcal{D}$ over $\mathcal{Y}$. Note that we do not assume the pixels in an image are i.i.d. The IoU for class $k$ over the data distribution is defined as:
\begin{equation}
\IoU_k(\theta) = \frac{p_k - p_{k0}(\theta)}{ p_k + p_{0k}(\theta)},
\end{equation}
where $p_{k0}(\theta)$ is the probability that a class $k$ pixel is observed and predicted as the background class by $\theta$, over the underlying data distribution $\mathcal{D}$. $p_{0k}(\theta)$ is similarly defined. We assume the empirical label distribution is an accurate estimation of the global label distribution $\mathcal{D}_{\mathcal{Y}}$, i.e.,
\begin{equation*}
p_k=P_{y\sim \mathcal{D}_{\mathcal{Y}}}(y=k)\approx p_{k,n} = \frac{1}{n}\sum\limits_{i=1}^{n}\mathbb{I}(y_i=k)
\end{equation*}
Similarly, the mIoU over the data distribution is defined as $\mIoU(\theta) = \frac{1}{c} \sum\limits_{k=1}^{c} \IoU_k(\theta)$. 

Ideally, a function $\theta$ should produce a high $\mIoU(\theta)$ to ensure the performance of $\theta$ on any data samples. Unfortunately, the data distribution $\mathcal{D}$ is usually fixed but unknown. Consequently, we can only optimize the empirical mIoU so that with a high probability it can lead to high $\mIoU(\theta)$. In the next section, we present our method to minimize the error bound between the empirical mIoU and $\mIoU(\theta)$ with a high probability, so the optimization of the empirical mIoU can also indicate the better $\mIoU(\theta)$.

\subsection{Theoretical motivation} \label{SEC:THEORY}
The mIoU is the average value of IoU scores over all classes, whereas in the medical image segmentation task, the label distributions are usually imbalanced. So equally treating all the pixels in training can lead to the biased IoU scores towards the majority classes. An intuitive solution is to set different margins for the pixel-samples belonging to different classes. Thus, we would derive an optimal margin setting for a smaller error bound between the empirical mIoU given by $\mIoU_n(\theta)$ and the expected mIoU given by $\mIoU(\theta)$.

Define the \textbf{margin} for $i$-th pixel in the image dataset with regard to the class $k$ as:
\begin{equation}\label{calmargin}
\lambda_{ik}=s_{ik}-\max_{j\neq k} s_{ij}.
\end{equation}
Similarly we can calculate the margins $\{\lambda_{ij}\}_{j=1}^{c}$ for pixel $i$ with regard to every class. If pixel $i$ is belong to the class $k$, then we would prefer a larger $\lambda_{ik}$ and a smaller $\lambda_{ij}, \forall j\neq k$, for a high confidence of prediction on the training dataset. 

We then combine the margin $\lambda_{ij}$  with a $\rho$-margin loss function $\phi_\rho(\cdot)$ defined in \cite[Definition 5.5]{mohri2018foundations}, to build the relationships between IoU score and the margin $\lambda_{ij}$. The $\rho$-margin loss is defined as:
\begin{equation}
\phi_{\rho}(\lambda)=\min \left(1, \max \left(0,1-\frac{\lambda}{\rho}\right)\right),
\end{equation}
which encourages the margin $\lambda$ to be larger than $\rho$ and provides an upper bound for 0-1 loss as illustrated in Figure \ref{lossfunplot}. We call the parameter $\rho$ \textbf{margin-offset}. We can then bound the empirical probabilities $p_{k0,n}(\theta)$ and $p_{0k,n}(\theta)$ in Eq.(\ref{emiou}) as: 
\begin{equation}\label{rhomloss}
\begin{aligned}
p_{k0,n}(\theta) & <\frac{1}{n}\sum\limits_{i\in Y_k} \phi_{\rho_{k0}}(\lambda_{ik}) = \ell_{k0,n}(\theta,\rho_{k0}), \\ 
p_{0k,n}(\theta) & <\frac{1}{n}\sum\limits_{i\in Y\setminus Y_k} \phi_{\rho_{0k}}(-\lambda_{ik}) = \ell_{0k,n}(\theta,\rho_{0k})
\end{aligned}
\end{equation} 
where we use $Y_k$ to denote the index set of pixels belong to class $k$ and $i\in Y\setminus Y_k$ to denote the index set of pixels excluding class $k$. $\rho_{0k}$ and $\rho_{k0}$ are pre-defined margin-offsets. Then, we can give a lower bound for Eq.(\ref{emiou}) as:
\begin{equation}\label{apploss}
\overline\IoU_{k,n}(\theta) = \frac{p_{k,n} - \ell_{k0,n}(\theta,\rho_{k0})}{ p_{k,n} + \ell_{0k,n}(\theta,\rho_{0k})},
\end{equation} 
and the related lower bound for $\mIoU_{n}(\theta)$: 
\begin{equation}
\overline\mIoU_{n}(\theta) = \frac{1}{c} \sum\limits_{k=1}^{c} \overline\IoU_{k,n}(\theta).
\end{equation}

We can then derive a generalization error bound regarding mIoU with the margin-offsets $\rho_{0k}$ and $\rho_{k0}$, based on the following theorem:
\begin{theorem}\label{miongen}
For any function $\theta\in\Theta$, define $\mu_k=\frac{\rho_{k0}}{\rho_{0k}}$ and $F=C(\Theta)+\sigma(\frac{1}{\eta})$. $C(\Theta)$ is some proper complexity measure of the hypothesis class $\Theta$ and $\sigma(\frac{1}{\eta}) \triangleq \frac{\rho_{\max}}{4c} \sqrt{2m\log \frac{2c}{\eta}}$ is typically a low-order term in $\frac{1}{\eta}$ with $\rho_{\max} = \max \{\rho_{i0},\rho_{0i}\}_{i=1}^c$. Given a training dataset of $n$ image pixels including $n_k$ pixels of class $k$, with each image consists of $m$ pixels, then for any $\eta>0$, with the probability at least $1-\eta$,
\begin{equation}
	\mIoU(\theta) \geq 
	\overline\mIoU_{n}(\theta)-\epsilon,
\end{equation}
where 
\begin{equation*}
\epsilon = \frac{1}{c}\sum\limits_{k=1}^{c} (\sqrt{n-n_k} + \frac{\sqrt{n_k}}{\mu_k})(\frac{n_k}{4cF}\rho_{0k}-\sqrt{n-n_k})^{-1}.
\end{equation*}
\end{theorem}

{\bf Proof.}We first give the proof that for each class $k$, the generalization error $\epsilon_k$ regarding $\overline\IoU_{k,n}(\theta)$, with probability $1-\frac{\eta}{c}$, satisfies the following inequality:
\begin{equation}\label{tobepf}
    \IoU_k(\theta) \geq \overline\IoU_{k,n}(\theta) - \epsilon_k.
\end{equation}
Averaging $\IoU_k(\theta)$, $\overline\IoU_{k,n}(\theta)$ and $\epsilon_k$ for $k = 1\cdots c$ and taking a union bound we can get the $\epsilon$ in Theorem 1.

With the definition of $\IoU_k(\theta)$, assume following inequality holds for non-negative $\epsilon_{0k}$ and $\epsilon_{k0}$:
\begin{equation} \label{ub}
\IoU_k(\theta) =  \frac{p_k - p_{k0}(\theta)}{ p_k + p_{0k}(\theta)} \geq 
\frac{p_k - (p_{k0}(\theta)-\epsilon_{k0})}{ p_k + (p_{0k}(\theta)-\epsilon_{0k})} - \epsilon_k.
\end{equation}
Solving above inequality, we can get:
\begin{equation} \label{finalepslk}
\epsilon_k = \frac{\frac{a_k}{b_k}\epsilon_{0k}+\epsilon_{k0}}{b_k-\epsilon_{0k}},
\end{equation}
where $a_k=p_k - p_{k0}(\theta)$ and $b_k = p_k + p_{0k}(\theta)$.

Next, we should get the values of $\epsilon_{0k}$ and $\epsilon_{k0}$ to satisfy following inequality:
\begin{equation}\label{ub2}
\frac{p_k - (p_{k0}(\theta)-\epsilon_{k0})}{ p_k + (p_{0k}(\theta)-\epsilon_{0k})} \geq \overline\IoU_{k,n}(\theta) = \frac{p_{k,n} - \ell_{k0,n}(\theta,\rho_{k0})}{ p_{k,n} + \ell_{0k,n}(\theta,\rho_{0k})},
\end{equation}
so we can simply substitute (\ref{ub2}) into (\ref{ub}) to complete the proof.

A sufficient condition for (\ref{ub2}) regarding $\epsilon_{0k}$ and $\epsilon_{k0}$ is:
\begin{equation}\label{eps2}
\begin{aligned}
p_{k0}(\theta)-\epsilon_{k0} &\leq \ell_{k0,n}(\theta,\rho_{k0}) = \frac{1}{n}\sum_{i\in Y_k} \phi_{\rho_{k0}}(\lambda_{ik})\\
p_{0k}(\theta)-\epsilon_{0k} &\leq \ell_{0k,n}(\theta,\rho_{0k}) = \frac{1}{n}\sum_{i\in Y\setminus Y_k} \phi_{\rho_{0k}}(-\lambda_{ik})
\end{aligned}
\end{equation}
Following the margin-based generalization bound in \cite[Theorem 9.2]{mohri2018foundations}, for the $n_k$ pixels belong to class $k$, with the probability at least $1-\frac{\eta}{2c}$, we have:
\begin{equation}\label{gb}
p_{k0}(\theta) - \ell_{k0,n}(\theta,\rho_{k0}) \leq \frac{n_k}{n}(\frac{4c}{\rho_{k0}}\mathfrak{R}_{n_k}(\Theta)+\sqrt{\frac{2m\log \frac{2c}{\eta}}{n_k}}),
\end{equation}
where $\mathfrak{R}_{n_k}(\Theta)$ is the Rademacher complexity for the hypothesis class $\Theta$ over the $n_k$ pixels belong to the foreground class $k$. Note that this inequality is slightly different from \cite[Theorem 9.2]{mohri2018foundations}, because the pixels are $m$-dependent for a dataset contains $m$-pixel images. We first apply the McDiarmid's inequality for $m$-dependent data \cite{liu2019mcdiarmid} to the proof of \cite[Theorem 3.3]{mohri2018foundations} to get a modified version of \cite[Theorem 3.3]{mohri2018foundations}. Then we use it in the proof of \cite[Theorem 9.2]{mohri2018foundations} to get the formulation of (\ref{gb}). 

The Rademacher complexity $\mathfrak{R}_{n_k}(\Theta)$ typically scales in $\sqrt{\frac{C(\Theta)}{n_k}}$ with $C(\Theta)$ being the some proper complexity measure of $\Theta$ \cite{neyshabur2018role}, and such a scale has also been used in related work (see \cite{cao2019learning} and the references therein). We can then rewrite (\ref{gb}) as:
\begin{equation}\label{epk0}
p_{k0}(\theta) - \ell_{k0,n}(\theta,\rho_{k0}) \leq \frac{\sqrt{n_k}}{n} \frac{4c}{\rho_{k0}}( C(\Theta) + \sigma(\frac{1}{\eta})  ),
\end{equation}
where $\sigma(\frac{1}{\eta}) \triangleq \frac{\rho_{\max}}{4c} \sqrt{2m\log \frac{2c}{\eta}}$ is typically a low-order term in $\frac{1}{\eta}$ with $\rho_{\max} = \max \{\rho_{i0},\rho_{0i}\}_{i=1}^c$. Similarly, let $F = C(\Theta) + \sigma(\frac{1}{\eta})$, with probability at least $1-\frac{\eta}{2c}$, we have:
\begin{equation}\label{ep0k}
p_{0k}(\theta) - \ell_{0k,n}(\theta,\rho_{0k}) \leq \frac{\sqrt{n-n_k}}{n}\frac{4c}{\rho_{0k}}  F,
\end{equation}
for the $n-n_k$ pixels that belong to the background class.

We then combine (\ref{epk0}), (\ref{ep0k}), (\ref{eps2}) and take a union bound over $\epsilon_{0k}$ and $\epsilon_{k0}$, to get following equations, with which (\ref{eps2}) holds with the probability at least $1-\frac{\eta}{c}$:
\begin{equation}
\begin{aligned}
\epsilon_{k0}&= \frac{\sqrt{n_k}}{n} \frac{4c}{\rho_{k0}}( C(\Theta) + \sigma(\frac{1}{\eta})  ) ,\\
\epsilon_{0k}&= \frac{\sqrt{n-n_k}}{n}\frac{4c}{\rho_{0k}} ( C(\Theta)+\sigma(\frac{1}{\eta}) ).
\end{aligned}
\end{equation}
Then we substitute above equations into (\ref{finalepslk}). Let $\mu_k = \frac{\rho_{k0}}{\rho_{0k}}$, we have:
\begin{equation}\label{bfapp}
\epsilon_k = \frac{\frac{a_k}{b_k}\sqrt{n-n_k} + \frac{\sqrt{n_k}}{\mu_k}}{\frac{b_k n}{4cF}\rho_{0k}-\sqrt{n-n_k}}
\end{equation} 
so that with the probability at least $1-\frac{\eta}{c}$ the inequality (\ref{tobepf}) holds. In practice, we do not know the values of $a_k$ and $b_k$ so that Eq.(\ref{bfapp}) has its own limitations. However, we know $\frac{a_k}{b_k}\leq 1$ and $b_k\geq p_k$ so we can get an even more useful bound:
\begin{equation}\label{aftapp}
\epsilon_k \leq \frac{ \sqrt{n-n_k} + \frac{\sqrt{n_k}}{\mu_k}}{\frac{n_k}{4cF}\rho_{0k}-\sqrt{n-n_k}}.
\end{equation}
Taking a union bound over all classes $k$, we can get the following inequality with the probability at least $1-\eta$:
\begin{equation} 
\mIoU(\theta) \geq \overline\mIoU_n(\theta) - \epsilon.
\end{equation}   
with 
\begin{equation} \label{finalresul}
\epsilon = \frac{1}{c}\sum_{k=1}^{c} \frac{ \sqrt{n-n_k} + \frac{\sqrt{n_k}}{\mu_k}}{\frac{n_k}{4cF}\rho_{0k}-\sqrt{n-n_k}},
\end{equation}  
where we complete the proof.

This theorem enables us to maximize the $\mIoU(\theta)$ on the data distribution by maximizing a lower bound $\overline\mIoU_{n}(\theta)$ for the empirical mIoU on the training dataset with a high probability. Meanwhile, we would prefer a small error bound $\epsilon$ so that the lower bound $\overline\mIoU_{n}(\theta)$ on the empirical mIoU could be a reliable estimation for $\mIoU(\theta)$. This scheme guarantees the performance of associated function $\theta$ on unseen data, e.g. the test data. 

Theorem \ref{miongen} also indicates that a smaller $\epsilon$ requires more pixels $n_k$ for each class, and a simpler fit function (for smaller $C(\Theta)$). Another important factor is that we can adjust the margin-offset $\rho_{0k}$ to minimize the error bound $\epsilon$. Note that increasing $\rho_{0i}$ also increases the $C(\Theta)$ implicitly, because a larger margin-offset may require more complex hypothesis class $\Theta$. Otherwise, $\overline\mIoU_{n}(\theta)$ decreases due to the under-fitting. Therefore, the scale of margin-offset should be tuned carefully. Besides, the direct calculation of the optimal margin-offsets in Theorem \ref{miongen} is difficult because it involves the complexity measure $C(\Theta)$, which is related to the structure of deep neural networks. Nevertheless, we can give the optimal proportions between $\rho_{0k}$'s that is irrelevant to $C(\Theta)$ by the following corollary:
\begin{corollary}\label{proprho}
Assume $\sum\limits_{k=1}^{c}\rho_{0k} = \text{some constant}.$ Let $\mu_k = \frac{p_{k}\sqrt{n_k}}{\upsilon(n-n_k)-p_{k}\sqrt{n-n_k}}$ with $\upsilon$ ($\upsilon>0$) being a hyper-parameter. Then the minimum of the error bound $\epsilon$ in Theorem \ref{miongen} is attained in the following condition:
\begin{equation}
 \frac{\rho_{0i}}{\rho_{0j}} = \frac{n_j}{n_i} \frac{\sqrt{n-n_i}}{\sqrt{n-n_j}},
\end{equation} 
\end{corollary}

{\bf Proof.} We substitute $\mu_k$ in (\ref{finalresul}) with $\frac{\sqrt{n_k}}{r(n/n_k-1)-\sqrt{n-n_k}}$, where $r$ is a hyper-parameter, we can get:
\begin{equation} 
\epsilon 
= \frac{1}{c}\sum_{k=1}^{c} \frac{\frac{r(n-n_k)}{n_k}}{\frac{n_k}{4cF}\rho_{0k}-\sqrt{n-n_k}}
= \frac{1}{c}\sum_{k=1}^{c} \frac{\frac{r(n-n_k)}{n_k^2}}{\frac{1}{4cF}\rho_{0k}-\frac{\sqrt{n-n_k}}{n_k}}.
\end{equation}
Let $x_k = \frac{r(n-n_k)}{n_k^2} $ and $y_k = \frac{1}{4cF}\rho_{0k}-\frac{\sqrt{n-n_k}}{n_k}$, according to Cauchy-Schwarz inequality we have:
\begin{equation}
\left(\sum_{k=1}^{c} \sqrt\frac{x_k}{y_k}\cdot \sqrt{y_k}\right)^2 \leq (\sum_{k=1}^{c}\frac{x_k}{y_k}) (\sum_{k=1}^{c} y_k),
\end{equation}
so that
\begin{equation}
\epsilon \geq \frac{1}{c}\cdot\frac{\left(\sum\limits_{k=1}^{c} \sqrt{x_k}\right)^2}{\sum\limits_{k=1}^{c} y_k} 
= \frac{r}{c}\cdot\frac{\left(\sum\limits_{k=1}^{c} \frac{\sqrt{n-n_k}}{n_k}\right)^2}{\frac{1}{4cF}\sum\limits_{k=1}^{c} \rho_{0k}- \sum\limits_{k=1}^{c}\frac{\sqrt{n-n_k}}{n_k}}. 
\end{equation}

The RHS of the equality is a constant because $r$ is a given hyper parameter and we assume  $\sum\limits_{k=1}^{c} \rho_{0k} = \text{some contant}$. The equality holds when $\frac{\sqrt{x_1}}{y_1}=...=\frac{\sqrt{x_c}}{y_c}$, which yields Corollary 1.1. 

Note that $\mu_k = \frac{\sqrt{n_k}}{r(n/n_k-1)-\sqrt{n-n_k}}$, while in Corollary 1.1 $\mu_k = \frac{p_k\sqrt{n_k}}{\upsilon(n-n_k)-p_k\sqrt{n-n_k}}$. These two conditions are essentially equivalent when $r$ and $\upsilon$ are hyper-parameters. To see this, simply let $r=n\upsilon$ and notice that $p_k=\frac{n_k}{n}$.

Corollary \ref{proprho} provides a theoretical guidance for setting the margin-offsets towards a smaller error bound $\epsilon$. The margin-offset $\rho_{0i}$ is proportional to $\frac{\sqrt{n-n_i}}{n_i}$, which indicates a larger margin is required for class $i$, with comparably fewer pixels. We introduce $\tau$ ($\tau>0$) to be the scale hyper-parameter of margin-offsets, which can be tuned on the validation dataset. A proper setting of $\tau$ and $\upsilon$ can provide a balance between $\epsilon$ and $\overline\mIoU_{n}(\theta)$ for the maximization of $\mIoU(\theta)$.
	
\subsection{A practical implementation}

The task of medical image segmentation is to maximize $\mIoU(\theta)$ for the best performance. Ideally, we should maximize its lower bound $\overline\mIoU_{n}(\theta)$ with a small error bound $\epsilon$. However, in the training of deep neural networks, the direct optimization of $\overline\mIoU_{n}(\theta)$ is impractical because the model is trained in a mini-batch manner. Unlike other decomposable evaluation metrics, such as classification accuracy, where the expectation of the metric on a mini-batch sample is equivalent to the metric on the whole dataset, the expectation of the mini-batch IoU is not equal to the overall IoU on the whole dataset. Accordingly, the lower bound $\overline\mIoU_{n}(\theta)$ has a similar problem. 

For practical implementation, we instead minimize the sum of each $\rho$-margin loss in $\overline\mIoU_{n}(\theta)$, with the optimal margin-offsets given in Corollary \ref{proprho}. By doing this, the empirical mIoU on the training dataset may be sub-optimal, but the margin-offsets can provide a guarantee for its generalization. So for a mini-batch of $n$ pixels, the loss $L(\theta)$ is calculated by:
\begin{equation}\label{mgloss}
\begin{aligned}
L(\theta) &= \sum\limits_{k=1}^{c} (\ell_{k0,n}(\theta,\rho_{k0})+\ell_{0k,n}(\theta,\rho_{0k}))\\
&= \frac{1}{n}\sum\limits_{k=1}^{c}\left(\sum\limits_{i\in Y_k} \phi_{\rho_{k0}}(\lambda_{ik})+\sum\limits_{i\in Y\setminus Y_k} \phi_{\rho_{0k}}(-\lambda_{ik})\right),
\end{aligned}
\end{equation}
with $\lambda_{ik}$ defined in Eq.(\ref{calmargin}). 

\begin{figure}[t]\center
	\includegraphics[width=0.4\textwidth]{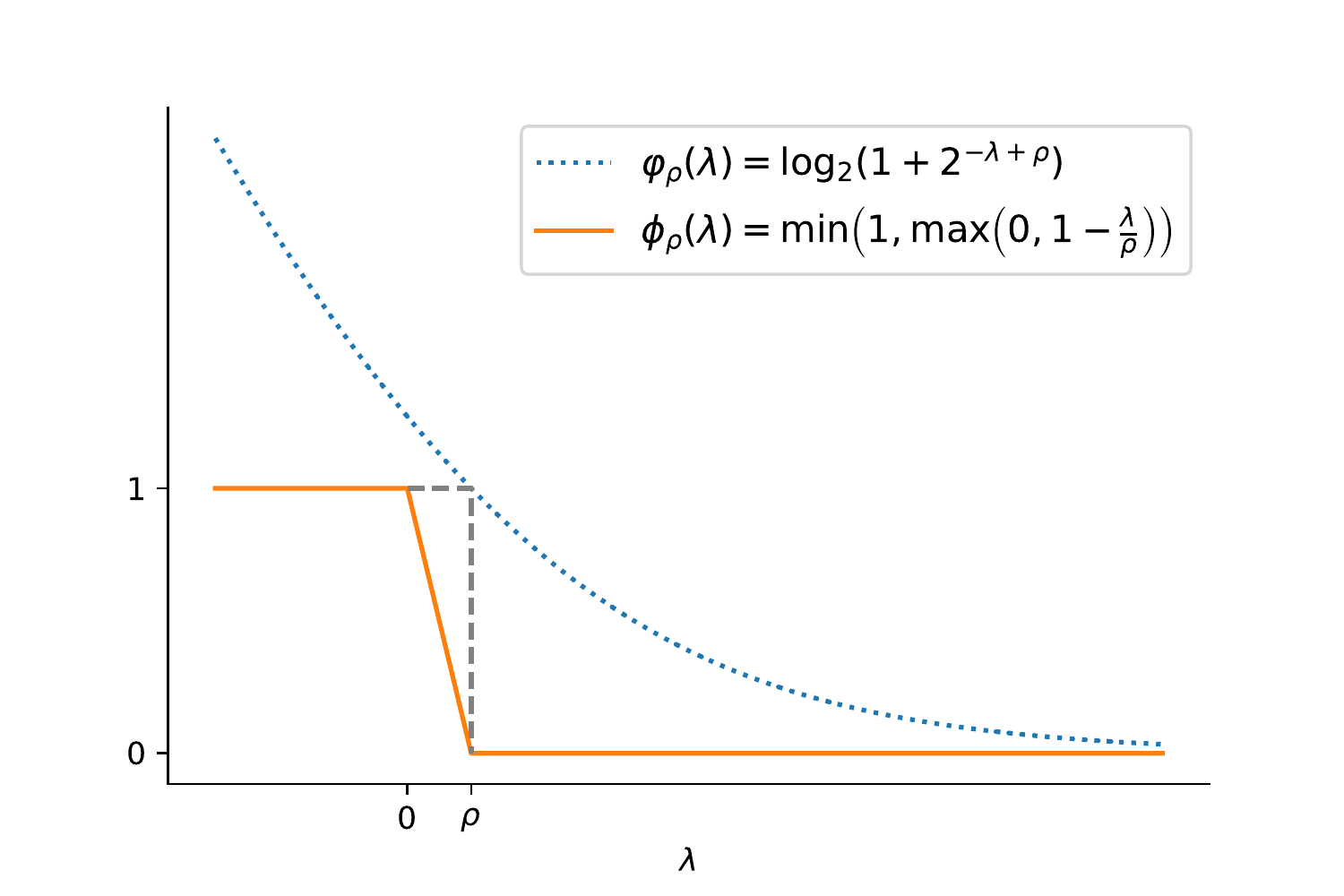}
	\caption{The $\rho$-calibrated log-loss (blue dotted line) and $\rho$-margin loss (orange solid line) functions. The $\rho$-margin loss is a upper bound for 0-1 loss. For the $\rho$-calibrated log-loss, $\varphi_\rho(\rho)=1$ and it upper bounds the $\rho$-margin loss.} \label{lossfunplot}
\end{figure}

In practice, the margin-offsets $\rho_{0k}$ and $\rho_{k0}$ may greatly influence the optimization of corresponding $\rho$-margin loss and bring instability in the optimization. Thus, we substitute the $\rho$-margin loss $\phi_\rho(\lambda)$ used in Eq.(\ref{mgloss}) with $\rho$-calibrated log-loss  $\varphi_\rho(\lambda)=\log_2(1+2^{-\lambda+\rho})$. The relationship between the $\rho$-margin loss $\phi_\rho(\lambda)$ and the $\rho$-calibrated log-loss $\varphi_\rho(\lambda)$ is illustrated in Figure \ref{lossfunplot}. As is shown in Figure \ref{lossfunplot}, the gradient regarding the $\rho$-margin loss can be prohibitively large when $\rho$ is very small, while the gradients outside the interval $(0,\rho)$ is zero. The $\rho$-calibrated log-loss bounds the $\rho$-margin loss from above and leads to:
\begin{equation}
\begin{aligned}
\ell_{k0,n}(\theta,\rho_{k0}) &< \frac{1}{n}\sum\limits_{i\in Y_k} \log_2(1+2^{-\lambda_{ik}+\rho_{k0}})=\underline\ell_{k0,n}(\theta,\rho_{k0}),
\end{aligned}
\end{equation}
and
\begin{equation}
\begin{aligned}
\ell_{0k,n}(\theta,\rho_{0k}) &< \frac{1}{n}\sum\limits_{i\in Y\setminus Y_k} \log_2(1+2^{\lambda_{ik}+\rho_{0k}})=\underline\ell_{0k,n}(\theta,\rho_{0k})
\end{aligned}
\end{equation}
Based on the above two inequalities, we simply use $\underline\ell_{k0,n}(\theta,\rho_{k0})$ and $\underline\ell_{0k,n}(\theta,\rho_{0k})$ to replace $\ell_{k0,n}(\theta,\rho_{k0})$ and $\ell_{0k,n}(\theta,\rho_{0k})$ in Eq.(\ref{mgloss}) as the final loss function.

\subsection{Complexity analysis}
%Our method requires $O(c)$ space complexity for storing margin loss parameters, which is negligible since the number of class $c$ is usually small compare to size of the segmentation model. 
%Given the output scores $s_{ij}\in\mathbb{R}^{n\times c}$ of $n$ pixels, computing the margin $\lambda_{ij}$ and the subsequent calibrated log-loss incurs $O(nc)$ time complexity. So compared to the cross-entropy loss, our margin calibration requires extra $O(nc)$ time overhead for the calculation of the margins. 
%Generally speaking, compared to the time complexity of calculating the output scores $s_{ij}$'s by a neural network, it is negligible.

Given the output scores $(s_{ij})\in\mathbb{R}^{n\times c}$ of $n$ pixels, the computation of the margin $\lambda_{ij}$ and the subsequent calibrated log-loss incurs $O(nc)$ time complexity. Specifically, compared to the cross-entropy loss, our calibration method requires extra $O(nc)$ time overhead to compute the margins.

\section{Experiments and analysis}

We use the recent proposed COPLE-Net \cite{TMI:COPLENET}, a variant of U-net, as the deep image segmentation architecture and compare the final segmented performance when applying commonly used learning objectives and our designed margin calibration method, respectively.

\subsection{Dataset}

We demonstrate the method on two publicly available medical image dataset: COVID-19 pneumonia CT scan (UESTC COVID-19 dataset \cite{TMI:COPLENET}) and Robotic Instrument segmentation \cite{ARXIV:ROBOTIC}. The COVID-19 dataset is collected from 10 different hospitals, in which the images have a large range of slice thickness/inter-slice spacing from 0.625mm to 8.0mm, and the pixel size ranges from 0.61mm to 0.93mm. The whole dataset contains two subsets, with 70 and 50 patient cases, respectively. The first subset ({\bf Part 1}) is coarse-labeled while the second one ({\bf Part 2}) is fine-labeled by experts. In the experiment, we used the fixed train/validation/test splits with 40/15/15 and 30/10/10 cases on the two subsets, respectively. The Robotic Instrument dataset provides $8\times225$-frame robotic surgical videos, where each part and type is manually annotated by a trained team. Here we conduct two segmentation tasks: {\bf Binary} instrument segmentation and {\bf Multi-class} instrument part segmentation. In the first task, each image is separated into da Vinci Xi instruments and the background class (ultrasound probe, surgical clips and porcine tissues). The second task is to correctly segment each articulating part of the instrument, including shaft, wrist, claspers and probe. In our experiment, this dataset is sequentially split into 1,200, 200 and 400 images according to the frame index for training, validation and testing, respectively. 

\subsection{Settings}

We implemented the segmentation model based on PyTorch. In the optimization, we employed the AdamW optimizer \cite{ICLR19:ADAMW} with the initial learning rate $10^{-4}$. We trained the COPLE-Net model with group normalization \cite{ECCV18:GN}, which allows setting a very small batch size to fit models in the limited GPU memory. Our experiments were conducted on a server equipped with an NVIDIA Titan X GPU card, and our implementation is publicly available at \url{https://github.com/XXX}.

\subsection{Results}

\subsubsection{Convergence study}

\begin{figure}[h]\centering\small
\begin{minipage}{0.22\textwidth}\centering
	\includegraphics[width=1\textwidth]{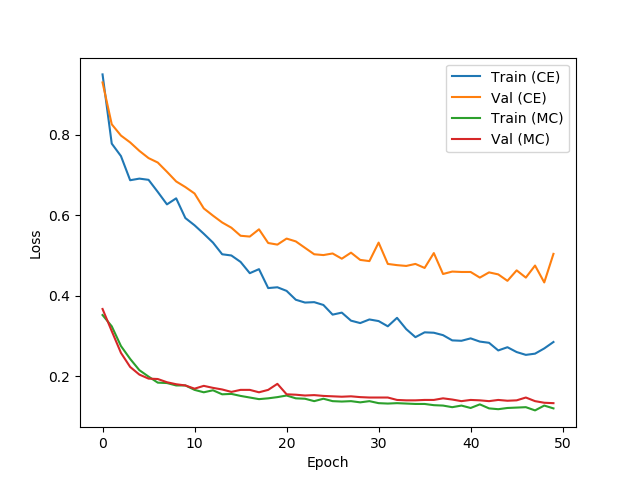}
	(a) Loss
\end{minipage}
\begin{minipage}{0.22\textwidth}\centering
	\includegraphics[width=1\textwidth]{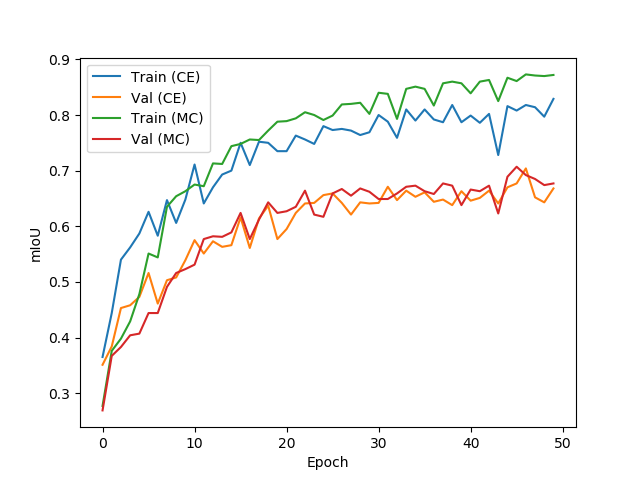}
	(b) mIoU
\end{minipage}
\caption{Loss and mIoU curves on the Robotic Instrument dataset (multi-class segmentation for different parts).}
\label{FIG:CONV}
\end{figure}

A very nice property of our proposed margin calibration method used in medical image segmentation is its tight correlation between empirical error and generalized error. We trained the segmentation model from the very beginning, using categorical cross-entropy (CE) and the proposed margin calibration (MC) in the first 50 optimization epochs, to record the loss values and mIoU scores, which are plotted in Figure \ref{FIG:CONV}. As we can see, the \emph{curve margin} between training loss and validation loss in cross-entropy gradually enlarges when training epoch increases, while the training loss and validation loss in our proposed margin calibration are generally close to each other. However, the absolute loss values using different loss functions have no direct correlation to the evaluation metric (mIoU in our case). From Figure \ref{FIG:CONV} (b) we can see using the margin calibration method, the mIoU has a comparable convergence speed to CE loss when applying the same optimization settings. Besides, our loss function leads to a higher training mIoU score, which can be accredited to the closer relationships between our loss function and mIoU score.
%From Figure \ref{FIG:CONV} (b) we can see using the margin calibration, the mIoU has a faster convergence when applying the same optimization settings.   

\subsubsection{Sensitivity analysis of hyper-parameters $\tau$ and $\upsilon$}
The hyper-parameters $\tau$ and $\upsilon$ control the scale of margins between the current foreground and the background classes. Proper margin-offsets can well assist the calibration of the pixel-class distribution variance. Thus, we set different values of $\tau$ and $\upsilon$ to observe the segmentation performance on the validation dataset. The loss values and mIoU scores are summarized in Table \ref{TB:CR}. We can observe that the settings of the two parameters do not significantly affect the actual performance of the models, which means the proposed margin calibration method is very robust to hyper-parameters.

\begin{table}[h]
\centering \small
\caption{Sensitivity of parameters $\tau$ and $\upsilon$ on the validation datasets. The results are reported based on the coarse-labelled binary segmentation of the COVID-19 dataset and the instrument parts segmentation (multi-class).}
\label{TB:CR}
\begin{tabular}{|c||cc|cc|}
\hline
\multirow{2}{*}{Parameter settings} & \multicolumn{2}{c|}{COVID-19} & \multicolumn{2}{c|}{Robotic Instrument}	 \\
\cline{2-5}
& Loss & IoU &Loss & mIoU  \\
\hline
$\tau=1,\upsilon=10$     & 0.034 & 72.8 & 0.086 & 71.2 \\
$\tau=1,\upsilon=1$     & 0.033 & 73.4 & 0.086 & 71.5 \\
$\tau=10,\upsilon=1$     & 0.031 & 73.1  & 0.093 & 72.7 \\
$\tau=10,\upsilon=0.1$     & 0.031 & 72.8 & 0.086 & 72.1 \\
$\tau=10,\upsilon=0.01$     & 0.033 & 72.7 & 0.090 & 71.8 \\
\hline
\end{tabular}
\end{table}

\subsubsection{Performance comparison using single learning objectives}

We tested the segmentation model with multiple learning objectives as baselines, including cross-entropy loss, generalized dice loss \cite{TMI:DICE}, focal loss \cite{CVPR17:FOCAL}, Tversky loss \cite{MLMIW17:TVERSKY}, lov{\'a}sz-softmax \cite{CVPR18:LOVASZ_SOFTMAX}. Cross-entropy is the most straight-forward loss function in classification, as medical image segmentation can be treated as a dense prediction for each image pixel. Although its learning objective is not so consistent with the evaluation matrics, cross-entropy is still a good loss function for early training due to its simple and fast computation. In our experiment, we used cross-entropy as the basic learning objective to pre-train the segmentation models for 50 epochs. After that, we applied different learning objectives independently to fine-tune the coarsely trained model. For fair comparisons, we did not use the CRF post-processing nor multi-scale prediction to bring complementary improvements. In model evaluation, we used per-pixel accuracy and IoU scores regarding different loss surrogates. 

We show the quantitative results of the two medical image datasets in Table \ref{TB:COVID-19} and \ref{TB:ROBOTIC}, respectively. The two evaluation metrics, pixel accuracy and IoU score, although have a very high correlation in terms of the absolute values, the best one single metric cannot guarantee the other. For example, simply using cross-entropy for the coarsely labelled COVID-19 pneumonia lesion segmentation task (Part 1 in Table \ref{TB:COVID-19}) achieve the best pixel accuracy, but its IoU score is not the best among the models with other loss functions. In fact, the IoU score is usually a better evaluation to quantify the percent overlap between the pixel-label output and target mask in medical image segmentation. Using the single loss function, our proposed margin calibration method obtains the best IoU or mIoU scores on the two datasets. Specifically, in the COVID-19 pneumonia lesion segmentation tasks, our method beats the second-best ones by 1.3\% and 0.4\% with the coarse- and fine-labelled CT image sets, respectively. The general performance on the fine-labelled set is much better due to the less noise. On the Robotic Instrument dataset, using the proposed margin calibration method as a single learning objective also outperforms other objective functions, with 1.4\% and 3.0\% performance boost in terms of IoU and mIoU scores for binary and multi-class segmentation, respectively. Also, although our method is not specifically designed to optimize the pixel accuracy, using the margin calibration can still achieve very promising performance. 

\begin{figure*}[t]\centering\small
\begin{minipage}{0.22\textwidth}\centering
	\includegraphics[width=1\textwidth]{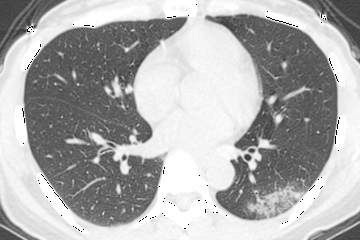}
	(a) Image
\end{minipage}
\begin{minipage}{0.22\textwidth}\centering
	\includegraphics[width=1\textwidth]{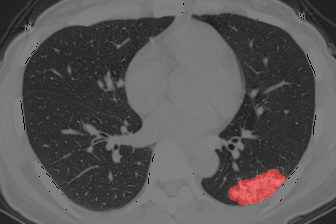}
	(b) Cross-entropy
\end{minipage}
\begin{minipage}{0.22\textwidth}\centering	
	\includegraphics[width=1\textwidth]{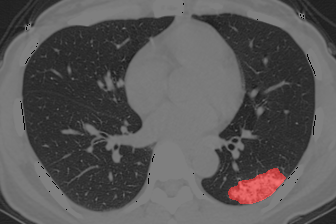} 
	(c) Dice loss   	
\end{minipage}
\begin{minipage}{0.22\textwidth}\centering
	\includegraphics[width=1\textwidth]{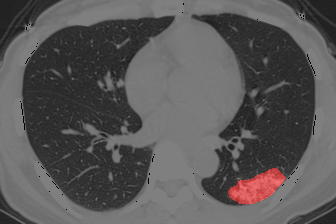}
	(d) Focal loss
\end{minipage}
\begin{minipage}{0.22\textwidth}\centering
	\includegraphics[width=1\textwidth]{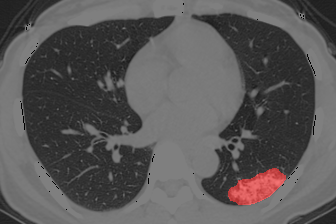}
	(e) Lov{\'a}sz-softmax 
\end{minipage}
\begin{minipage}{0.22\textwidth}\centering
	\includegraphics[width=1\textwidth]{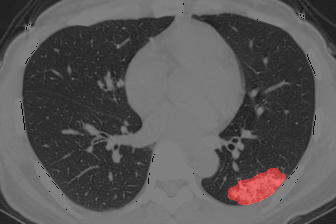}
	(f) Tversky loss
\end{minipage}
\begin{minipage}{0.22\textwidth}\centering
	\includegraphics[width=1\textwidth]{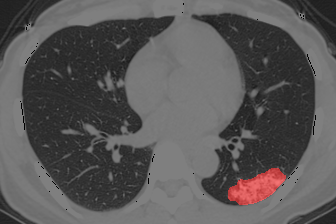}
	(g) Margin calibration (ours)
\end{minipage}
\begin{minipage}{0.22\textwidth}\centering
	\includegraphics[width=1\textwidth]{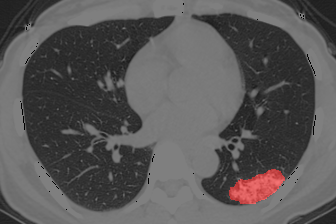}
	(h) Ground truth
\end{minipage}
\caption{Segmentation examples on COVID-19 test set (fine-labeled).}
\label{FIG:COVID19}
\end{figure*}

\begin{figure*}[t]\centering\small
\begin{minipage}{0.22\textwidth}\centering
	\includegraphics[width=1\textwidth]{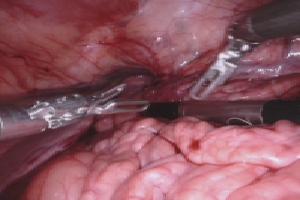}
	(a) Image
\end{minipage}
\begin{minipage}{0.22\textwidth}\centering
	\includegraphics[width=1\textwidth]{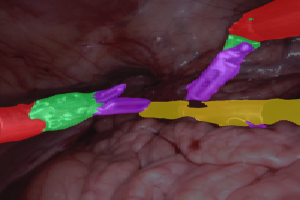}
	(b) Cross-entropy
\end{minipage}
\begin{minipage}{0.22\textwidth}\centering	
	\includegraphics[width=1\textwidth]{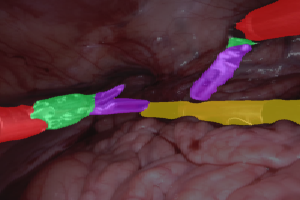} 
	(c) Dice loss   	
\end{minipage}
\begin{minipage}{0.22\textwidth}\centering
	\includegraphics[width=1\textwidth]{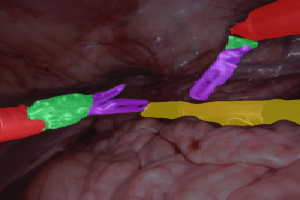}
	(d) Focal loss
\end{minipage}
\begin{minipage}{0.22\textwidth}\centering
	\includegraphics[width=1\textwidth]{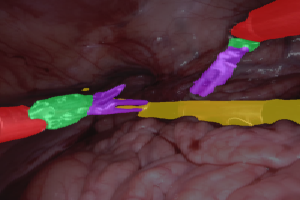}
	(e) Lov{\'a}sz-softmax 
\end{minipage}
\begin{minipage}{0.22\textwidth}\centering
	\includegraphics[width=1\textwidth]{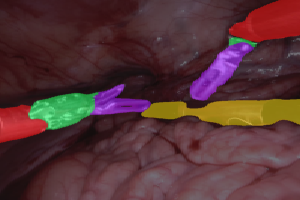}
	(f) Tversky loss
\end{minipage}
\begin{minipage}{0.22\textwidth}\centering
	\includegraphics[width=1\textwidth]{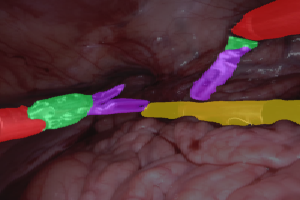}
	(g) Margin calibration (ours)
\end{minipage}
\begin{minipage}{0.22\textwidth}\centering
	\includegraphics[width=1\textwidth]{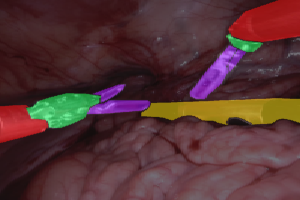}
	(h) Ground truth
\end{minipage}
\caption{Segmentation examples on Robotic Instrument test set.}
\label{FIG:ROBOTIC}
\end{figure*}

We illustrate the segmentation examples in Figure \ref{FIG:COVID19} and Figure \ref{FIG:ROBOTIC} on the two datasets, respectively. By observing the results of COVID-19 pneumonia lesion, using different learning objectives in the COPLE-Net obtains very similar results, thus we cannot see obvious differences. On the visualization of the multi-class segmentation on the Robotic Instrument dataset, we can see that applying the proposed method, different parts can be better segmented, forming more smooth contours and obtaining more accurate results.

\begin{table}[h]
\centering \small
\caption{Segmentation performance comparison on the COVID-19 test set.}
\label{TB:COVID-19}
\begin{tabular}{|c||cc|cc|}
\hline
\multirow{2}{*}{Method} & \multicolumn{2}{c|}{Part 1} & \multicolumn{2}{c|}{Part 2}	 \\
\cline{2-5}
 & Pixel Acc.  & IoU & Pixel Acc.  & IoU  \\
\hline
Cross-entropy & 81.1 & 69.6 & 86.6 & 76.8 \\
Dice loss     &  80.7 & 70.0 & 85.5 & 76.1 \\
Tversky loss  & 80.7  & 68.1 & {\bf 89.1} & 76.0 \\
Focal loss    & 80.0 & 68.5 & 87.3 & 76.9 \\
lov{\'a}sz-softmax & 79.6 & 70.0 & 88.2 & 77.7 \\
Ours & {\bf 83.8} & {\bf 71.3}  & 88.3 & {\bf 78.1} \\
\hline 
\end{tabular}
\end{table}

\begin{table}[h]
\centering \small
\caption{Segmentation performance comparison on the Robotic Instrument test set.}
\label{TB:ROBOTIC}
\begin{tabular}{|c||cc|cc|}
\hline
\multirow{2}{*}{Method} & \multicolumn{2}{c|}{Binary} & \multicolumn{2}{c|}{Multi-class}	 \\
\cline{2-5}
 & Pixel Acc.  & IoU & Pixel Acc.  & mIoU  \\
\hline
Cross-entropy     & 94.6 & 86.0 & 81.1 & 66.2 \\
Dice loss    & {\bf 97.7} & 84.7 & 78.5 & 67.6 \\
Tversky loss & 96.4 & 85.9 & 79.7 & 68.6 \\
Focal loss  & 94.8 & 86.2 & 80.0 & 69.5 \\
lov{\'a}sz-softmax & 96.3 & 86.0 & 80.1 & 68.9 \\
Ours & 95.8 & {\bf 87.4} & {\bf 81.5} & {\bf 72.5} \\
\hline
\end{tabular}
\end{table}

\begin{table}[h]
\centering \small
\caption{Per-class IoU scores for multi-class segmentation on the Robotic instrument test set.}
\label{TB:ROBOTIC_PERCLASS}
\begin{tabular}{|c||cccc|}
\hline
Method & Shaft & Wrist & Claspers & Probe \\
\hline
Cross-entropy & 81.1 & 55.3 & 55.5 & 72.9 \\
Dice loss    & 83.8 & 61.8 & 54.2 & 70.5 \\
Tversky loss & 84.9 & 64.8 & 55.4 & 69.3 \\
Focal loss   & 86.5 & 62.9 & 56.5 & 72.0 \\
lov{\'a}sz-softmax & 86.3 & 64.4 & 55.5 & 69.3 \\
Ours & {\bf 88.2} & {\bf 67.1} & {\bf 61.1} & {\bf 73.4} \\
\hline
\end{tabular}
\end{table}

\subsubsection{Performance using loss function combinations}

Applying multiple loss functions simultaneously in training an image segmentation model is a common practice. Among the baselines, dice loss and Tversky loss are region-based loss functions, while Lov{\'a}sz-softmax and focal loss, as well as our proposed margin calibration, focus more on data-distribution. So we simply used our method in conjunction with dice loss and Tversky loss as the learning objectives to train the COPLE-Net models. On the two datasets, the IoU scores can be further boosted in general (see
Table \ref{TB:COMPOUND}). 

\begin{table}[h]
\centering \small
\caption{Segmentation performance (IoU \& mIoU) using compound learning objectives.}
\label{TB:COMPOUND}
\begin{tabular}{|c||cc|cc|}
\hline
\multirow{2}{*}{Method} & \multicolumn{2}{c|}{COVID-19} & \multicolumn{2}{c|}{Robotic Instrument}	\\
\cline{2-5}
& Part 1 & Part 2 & Binary & Multi-class  \\
\hline
Ours + Tversky loss  & 71.6 & 78.5 & 87.7 & 73.1 \\
Ours + Dice loss     & 71.5 & 78.0 & 87.8 & 73.2 \\
\hline
\end{tabular}
\end{table}

\section{Conclusion}

We have presented a versatile margin calibration method for a better learning objective to optimize the Jaccard index in medical image segmentation. With the consideration of both empirical performance and the error bound regarding the generalization performance, the scheme can increase the discriminative power with a better generalization ability. We gave both theoretical and experimental analysis to demonstrate its effectiveness, substantially improving the IoU scores by inserting it into a deep learning-based medical image segmentation model.

\bibliography{reference.bib}

\end{document}